\documentclass{article}
\usepackage[utf8]{inputenc}

\usepackage{amsmath}
\usepackage{amssymb}
\usepackage{authblk}
\usepackage{booktabs} 
\usepackage{makecell}
\usepackage{multirow}
\usepackage{pdfpages}

\title{Searching Intrinsic Dimensions of Vision Transformers}
\author{Fanghui Xue}
\author{Biao Yang}
\author{Yingyong Qi}
\author{Jack Xin}

\affil{Department of Mathematics\\ University of California, Irvine\\ Irvine, CA 92697, USA}

\begin{document}

\maketitle

\begin{abstract}
    It has been shown by many researchers that transformers perform as well as convolutional neural networks in many computer vision tasks. Meanwhile, the large computational costs of its attention module hinder further studies and applications on edge devices. Some pruning methods have been developed to construct efficient vision transformers, but most of them have considered image classification tasks only. Inspired by these results, we propose SiDT, a method for pruning vision transformer backbones on more complicated vision tasks like object detection, based on the search of transformer dimensions. Experiments on CIFAR-100 and COCO datasets show that the backbones with 20\% or 40\% dimensions/parameters pruned can have similar or even better performance than the unpruned models. Moreover, we have also provided the complexity analysis and comparisons with the previous pruning methods.
\end{abstract}
\makeatletter{\renewcommand*{\@makefnmark}{}
\footnotetext{To appear in the PICIES-22 conference.}\makeatother}

\section{Introduction}
Unlike the convolutional or recurrent neural networks (CNN or RNN) \cite{Goodfellow-et-al-2016}, 
{\it transformers} are the models based completely or partially on the {\it attention} mechanisms. They are originally proposed to learn global dependency for sequence transduction tasks \cite{vaswani2017attention}, and have obtained better performance and training efficiency. Besides its success in language models, transformers have also been widely studied in computer vision tasks. One of the directions is to replace the CNN backbones by transformers. In other words, transformers are used to extract features from images, and the features are processed by various heads to solve various tasks after that. Among these transformer based models, ViT \cite{dosovitskiy2020image}, DeiT \cite{touvron2021training} and Swin Transformer \cite{liu2021swin} have achieved high performance in multiple tasks like image classification, object detection, segmentation, etc.

The general architecture of the transformer for sequence modeling is composed of an encoder module and a subsequent decoder module. The encoder module is a stack of a few sequential encoder blocks, with each of them containing a {\it self-attention} (SA) layer and a fully connected {\it feed-forward network}, with a residual structure \cite{he2016deep}, and a layernorm applied after the summation of the shortcut and the residual. While the feed-forward network consists simply of two fully connected layers, the self-attention layer is computed through a {\it multi-head attention} (MSA) mechanism \cite{vaswani2017attention}, which is more complicated and usually requires more computational resources than the convolution operations used in CNNs. Therefore, pruning methods \cite{zhu2021vision, yang2021nvit, yu2022width} have been proposed to construct efficient vision transformers. However, most of them only consider pruning DeiT on the image classification task. In this paper, we present a pruning method for transformer backbone which is valid on both image classification and object detection tasks. Since our method aims to search for the intrinsic dimensions (i.e., the possible lowest dimensions to maintain network performance) of transformers, we name it SiDT in the rest of the paper. Although SiDT is inspired by previous pruning methods like Network Slimming \cite{liu2017learning} and Vision Transformer Pruning (VTP) \cite{zhu2021vision}, it has its own merits:
\begin{itemize}
\item •	SiDT can prune transformers for not only classification tasks, but also other vision tasks like object detection.
\item We have analyzed the computational complexity of the unpruned and the pruned models.
\item The models with 20\% or 40\% dimensions pruned perform similarly or even better than the unpruned model.
\item SiDT prunes the dimensions of linear embeddings, different from the feature pruning of VTP.
\end{itemize}

\section{Related Work}
\subsection{Vision Transformers}
Vision Transformer (ViT) \cite{dosovitskiy2020image} is among the vision models whose backbones are purely transformers. ViT has partitioned the input image into small patches to mimic the tokens in the language transformers. Instead of pixels, these patches are embedded into features of certain dimensions, serving as the input of the attention module. Since its job is to learn representations, ViT has included the encoder module only, i.e., a stack of multi-head self-attentions. Despite ViT's high accuracy on image classification, there are some concerns about its quadratic computational complexity on the number of queries $n$. That means the complexity is also quadratic on the input resolution $H\times W$, whereas the convolution operation has linear complexity. ViT has also been restricted to image classification, since pixel-level tasks like segmentation typically need to deal with high resolution features. 

A window-based transformer called Swin Transformer \cite{liu2021swin} has then been proposed for these more complicated vision tasks. Similar to ViT, Swin has also provided a series of backbones which are based purely on transformers, especially the transformer encoders. The first advantage of Swin is that it can generate hierarchical features so that they can be used to solve semantic segmentation and object detection tasks with suitable heads. To obtain features of different resolutions, Swin has merged $2\times 2=4$ image patches into 1 patch at the end of each architecture stage. Since the size of patches is fixed, the image height and the width are both reduced by a half after merging. The overall transformer architecture is divided into one initial stage without merging and three intermediate stages with merging, and hence it can produce features of four resolution levels. Another advantage comes from the window-based multi-head self-attention (W-MSA) with shifting. Compared with the quadratic complexity of MSA, W-MSA has achieved a linear complexity from computing the attentions locally, within a small window of patches. Global information across different windows is then exchanged via shifting the window partitions. 

\subsection{Dimension Pruning}\label{formulation}
The dimension/channel pruning problem of CNNs can be solved by adding group sparsity to the convolutional weights  \cite{yuan2006model, wen2016learning}, or formulated as a neural architecture search problem  \cite{zoph2018learning,liu2019darts}. Among them, a method called Network Slimming (NetSlim) has been proposed based on learning the channel scaling factors \cite{liu2017learning}, which is able to reduce the model complexity and computational cost, and preserve the accuracy at the same time. These channel scaling factors are simply defined to be the learnable scale parameters $\gamma$ of the batch normalization layer, and the channels corresponding to low scales are pruned. To learn sparse scales, the $\ell_1$ regularization of these scale parameters is added to the loss during training. After being trained with $\ell_1$ sparsity and the channels with low scales pruned, the model is further fine-tuned to achieve better performance. We shall be aware that the regularization term is not added to the convolutional weights, but directly to the scale parameters, which play a similar role as the architecture parameters in the differentiable neural architecture search context  \cite{liu2019darts}. That is why searching for dimensions is indeed dimension pruning.

Similar to the channel pruning in CNNs, there are also some studies for vision transformer pruning. Inspired by NetSlim, VTP \cite{zhu2021vision} has assigned scoring parameters to the features before the linear embedding or projection layers and pruned the dimensions of these features which are corresponding to low scores. Since the dimension of the linear layers depend on the dimension of the input features, the parameters of these layers are also reduced. Another pruning method has been proposed in NViT \cite{yang2021nvit}, which is based on the scores of grouped  structural parameters. The scores are different from those of VTP as they are computed directly from the weight parameters. NViT has taken pruning the number of heads and the latency on hardware into account. Moreover, it has been pointed out that having the same dimensions across all layers in the conventional transformer design might not be optimal \cite{yang2021nvit}, which inspires the studies of automated transformer architecture design. 

\begin{figure}
      \centering
      {\includegraphics[width=1\textwidth]{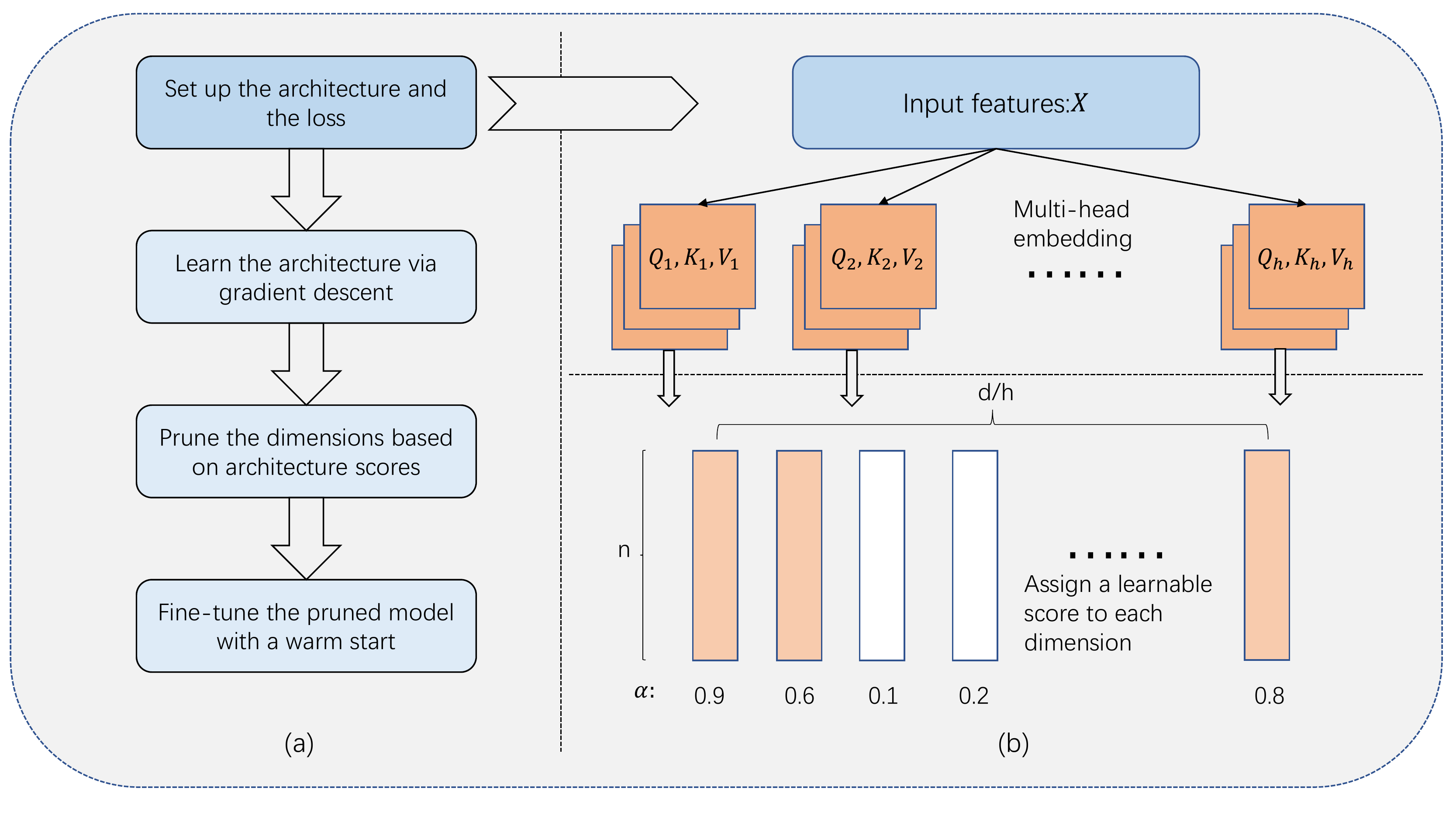}}
    \caption{(a) The stages of transformer pruning. (b) Assign the scoring matrix $\textbf{\emph{A}} = \mathrm{diag}(\alpha)$ to the output dimensions of multi-head queries, keys and values. } 
      \label{diag}
    \end{figure}

These pruning methods have obtained high pruning ratio with a very small accuracy loss for vision transformers like DeiT \cite{touvron2021training}, on the image classification tasks. It would be natural to consider pruning Swin or other light transformer backbones for multiple computer vision tasks. WDPruning \cite{yu2022width} is a direct pruning method for Swin on ImageNet classification, without the fine-tuning stage. It has also provided an option for depth pruning, and an automated learned pruning ratio based on learnable thresholds of saliency scores. However, experimental results has shown worse accuracy of the pruned models, as it has not been fine-tuned. Inspired by these previous studies, we consider pruning Swin backbone as dimension search in this paper. Before we specify the details of each stage, we summarize a general framework for searching the dimensions of operations \cite{liu2017learning,zhu2021vision} (see also Fig. \ref{diag}(a)):
\begin{itemize}
\item Specify the architecture parameters for representing the dimensions of the operations.
\item Set up a loss function which involves the architecture parameters and the other learnable parameters.
\item Optimize the loss via gradient descent and prune the network based on the values of the architecture parameters.
\item Fine-tune the pruned network.
\end{itemize}

\section{Method}
\textbf{Architecture parameters.} For the dimension search of transformers, we still follow the four stages summarized in Section \ref{formulation}. Since the searching, pruning and fine-tuning stages are similar, the key difference is how we set up the architecture parameters. Whereas we prune convolution operations in CNNs, there are a few types of operations for different transformers. So we discuss in detail the strategies of setting up architectures parameters for MSA, W-MSA and {\it multilayer perceptron} (MLP) \cite{liu2021swin}. Suppose again the batch size is $N=1$, $\textbf{\emph{X}}\in\mathbb{R}^{d\times H\times W}$ is the input feature map with $H$ and $W$ the resolution and $d$ the dimension of the feature. Set $n=H\times W$, we obtain the transformed input feature $\textbf{\emph{X}}\in\mathbb{R}^{n\times d}$. 

For SA \cite{vaswani2017attention}, $\textbf{\emph{X}}$ is linearly embedded into the query $\textbf{\emph{Q}}$, key $\textbf{\emph{K}}$ and value $\textbf{\emph{V}}$ of the same shapes:
\begin{align*}
    \textbf{\emph{Q}} =  \textbf{\emph{X}}\textbf{\emph{W}}_Q,\,\, \textbf{\emph{K}} = \textbf{\emph{X}}\textbf{\emph{W}}_K,\,\,
    \textbf{\emph{V}} = \textbf{\emph{X}}\textbf{\emph{W}}_V,
\end{align*}
where the embedding matrices $\textbf{\emph{W}}_Q, \textbf{\emph{W}}_K, \textbf{\emph{W}}_V \in \mathbb{R}^{ d\times d}$, if the embedding dimensions for the query, key and value are also equal to $d$. Then the attention map $a$ is computed via the softmax function $\sigma$ of the scaled product of the query and the key:
\begin{align*}
    a(\textbf{\emph{Q}}, \textbf{\emph{K}}) = \sigma(\textbf{\emph{Q}}\textbf{\emph{K}}^{T}/\sqrt{d}) \in \mathbb{R}^{n\times n},
\end{align*}
and assigned to the value to compute the output of SA:
\begin{align*}
    SA(\textbf{\emph{Q}}, \textbf{\emph{K}}, \textbf{\emph{V}}) = \sigma(\textbf{\emph{Q}}\textbf{\emph{K}}^{T}/\sqrt{d})\textbf{\emph{V}} \in \mathbb{R}^{n\times d}.
\end{align*}
Note that the output of SA has the same shape as the input $\textbf{\emph{X}}$. To set up the architecture parameters, we apply a uniform score matrix $\textbf{\emph{A}}$ for $\textbf{\emph{Q}}$,  $\textbf{\emph{K}}$ and $\textbf{\emph{V}}$ via matrix multiplication:
\begin{align*}
    \widetilde{\textbf{\emph{Q}}} =  \textbf{\emph{Q}}\textbf{\emph{A}},\,\, \widetilde{\textbf{\emph{K}}} =  \textbf{\emph{K}}\textbf{\emph{A}},\,\,
    \widetilde{\textbf{\emph{V}}} =  \textbf{\emph{V}}\textbf{\emph{A}},
\end{align*}
where $\textbf{\emph{A}} \in\mathbb{R}^{d\times d}$ is a diagonal matrix whose diagonal elements are the architecture parameters $\alpha_i$ for $i=1,2,...,d$. That is to say, we assign a score $\alpha_i$ to the $i$-th dimension of the $d$-dimensional query, and also to the key and value at the same $i$-th dimension. Then we compute the SA module based on the scored query, key and value, and obtain $SA(\widetilde{\textbf{\emph{Q}}}, \widetilde{\textbf{\emph{K}}}, \widetilde{\textbf{\emph{V}})}$.

For MSA, we need to compute multiple SA modules and each of them is a {\it head}. Let $h$ be the number of heads. For $j=1,...,h$, we also compute $\textbf{\emph{Q}}_j$,  $\textbf{\emph{K}}_j$ and $\textbf{\emph{V}}_j\in \mathbb{R}^{n\times d/h}$ through linear embedding of $\textbf{\emph{X}}$ via $\textbf{\emph{W}}_{Q,j}$, $\textbf{\emph{W}}_{K,j}$ and $\textbf{\emph{W}}_{V,j}\in \mathbb{R}^{d\times d/h}$ like that of SA, and obtain the heads:
\begin{align*}
    \textbf{\emph{H}}_j = SA(\textbf{\emph{Q}}_j, \textbf{\emph{K}}_j, \textbf{\emph{V}}_j)\in \mathbb{R}^{n\times d/h}.
\end{align*}
With $\textbf{\emph{Q}}$,  $\textbf{\emph{K}}$ and $\textbf{\emph{V}}$ the concatenations of $\textbf{\emph{Q}}_j$,  $\textbf{\emph{K}}_j$ and $\textbf{\emph{V}}_j$, the output of the MSA module is computed by concatenating the heads and projecting linearly via $\textbf{\emph{W}}_O\in \mathbb{R}^{d\times d}$:
\begin{align*}
    MSA(\textbf{\emph{Q}}, \textbf{\emph{K}}, \textbf{\emph{V}}) = [\textbf{\emph{H}}_1, \textbf{\emph{H}}_2, ..., \textbf{\emph{H}}_h]\textbf{\emph{W}}_O \in \mathbb{R}^{n\times d}.
\end{align*}
We use a stronger scoring matrix $\textbf{\emph{A}}\in \mathbb{R}^{d/h\times d/h}$ for MSA, which is not only uniform over the query, key and value, but also over all the heads:
\begin{align*}
    \widetilde{\textbf{\emph{Q}}}_j =  \textbf{\emph{Q}}_j\textbf{\emph{A}},\,\, \widetilde{\textbf{\emph{K}}}_j =  \textbf{\emph{K}}_j\textbf{\emph{A}},\,\,
    \widetilde{\textbf{\emph{V}}}_j =  \textbf{\emph{V}}_j\textbf{\emph{A}},
\end{align*}
for $j=1,2,...,h$. Then we compute the new MSA module and obtain $\widetilde{\textbf{\emph{H}}}_j = SA(\widetilde{\textbf{\emph{Q}}}_j, \widetilde{\textbf{\emph{K}}}_j, \widetilde{\textbf{\emph{V}}}_j)$ and:
\begin{align*}
    MSA(\widetilde{\textbf{\emph{Q}}}, \widetilde{\textbf{\emph{K}}}, \widetilde{\textbf{\emph{V}}}) = [\widetilde{\textbf{\emph{H}}}_1, \widetilde{\textbf{\emph{H}}}_2, ..., \widetilde{\textbf{\emph{H}}}_h]\textbf{\emph{W}}_O .
\end{align*}
For W-MSA, the input features $\textbf{\emph{X}}\in\mathbb{R}^{n\times d}$ are divided into a few windows of size $M\times M$, and MSA is computed locally within these windows. That is to say, we reshape $\textbf{\emph{X}}$ to be a tensor in $\mathbb{R}^{n/M^2 \times M^2 \times d}$, and obtain $\textbf{\emph{Q}}_j$,  $\textbf{\emph{K}}_j$ and $\textbf{\emph{V}}_j\in \mathbb{R}^{n/M^2 \times M^2 \times d/h}$ for $j=1,2,...,h$ after embedding of multi-head. Here $\textbf{\emph{Q}}_j$,  $\textbf{\emph{K}}_j$ and $\textbf{\emph{V}}_j$ can be viewed as the concatenations of $\textbf{\emph{Q}}_{j,l}$,  $\textbf{\emph{K}}_{j,l}$ and $\textbf{\emph{V}}_{j,l}\in\mathbb{R}^{M^2 \times d/h}$ for $l=1,2,...,n/M^2$. For each window, we compute the MSA module and obtain $\textbf{\emph{W}}_{,l} = MSA(\textbf{\emph{Q}}_{,l}, \textbf{\emph{K}}_{,l}, \textbf{\emph{V}}_{,l})\in \mathbb{R}^{M^2\times d}$. Finally, we rearrange the outputs of these windows and obtain:
\begin{align*}
    W\text{-}MSA(\textbf{\emph{Q}}, \textbf{\emph{K}}, \textbf{\emph{V}}) = [\textbf{\emph{W}}_{,1}, \textbf{\emph{W}}_{,2}, ..., \textbf{\emph{W}}_{,n/M^2}] \in \mathbb{R}^{n\times d}.
\end{align*}
To set up the architecture parameters for W-MSA, again we use a uniform {\it scoring matrix} $\textbf{\emph{A}}\in \mathbb{R}^{d/h\times d/h}$ for the query, key and value, over all the heads and windows:
\begin{align*}
    \widetilde{\textbf{\emph{Q}}}_{j,l} =  \textbf{\emph{Q}}_{j,l}\textbf{\emph{A}},\,\, \widetilde{\textbf{\emph{K}}}_{j,l} =  \textbf{\emph{K}}_{j,l}\textbf{\emph{A}},\,\,
    \widetilde{\textbf{\emph{V}}}_{j,l} =  \textbf{\emph{V}}_{j,l}\textbf{\emph{A}}.
\end{align*}
Then we have $\widetilde{\textbf{\emph{W}}}_{,l} = MSA(\widetilde{\textbf{\emph{Q}}}_{,l}, \widetilde{\textbf{\emph{K}}}_{,l}, \widetilde{\textbf{\emph{V}}}_{,l})$ and 
\begin{align*}
    W\text{-}MSA(\widetilde{\textbf{\emph{Q}}}, \widetilde{\textbf{\emph{K}}}, \widetilde{\textbf{\emph{V}}}) = [\widetilde{\textbf{\emph{W}}}_{,1}, \widetilde{\textbf{\emph{W}}}_{,2}, ..., \widetilde{\textbf{\emph{W}}}_{,n/M^2}].
\end{align*}

The last module to be discussed is MLP \cite{liu2021swin}, which simply contains two linear layers with activation. Suppose $\textbf{\emph{X}}\in\mathbb{R}^{n\times d}$ is the input feature, and $d_m$ represents the dimensions of the hidden state. Suppose further $\textbf{\emph{W}}_1\in\mathbb{R}^{d\times d_m}$ and $\textbf{\emph{W}}_2\in\mathbb{R}^{d_m\times d}$ are two matrices for linear embedding, $\sigma_{MLP}$ is the activation. Then we have:
\begin{align*}
    MLP(\textbf{\emph{X}}) = \sigma_{MLP}(\textbf{\emph{X}}\textbf{\emph{W}}_1)\textbf{\emph{W}}_2 \in \mathbb{R}^{n\times d}.
\end{align*}
The scoring matrix $\textbf{\emph{A}}$ is applied immediately after $\textbf{\emph{W}}_1$ through matrix multiplication, and get $\sigma_{MLP}(\textbf{\emph{X}}\textbf{\emph{W}}_1\textbf{\emph{A}})\textbf{\emph{W}}_2$. Here $\textbf{\emph{A}}$ can be viewed as the scores for the dimensions of the hidden state.

\textbf{Pruning.} The four-stage pruning procedure is summarized in Fig. \ref{diag}. During the searching stage, the elements in the scoring matrix $\textbf{\emph{A}}$
are regularized by $\ell_1$ norm like NetSlim \cite{liu2017learning} , and involved in the overall loss:
\begin{align*}
    L=l(\textbf{\emph{X}},\textbf{\emph{T}};\textbf{\emph{W}}) + \gamma\ell_1(\textbf{\emph{A}}),
\end{align*}
where $l$ is the classification or detection loss, $l_1$ is the $\ell_1$ loss,  $\textbf{\emph{X}}$, $\textbf{\emph{T}}$ and $\textbf{\emph{A}}$ are the input, target and the architecture parameters, and $\textbf{\emph{W}}$ represents the other learnable parameters. $\gamma$ is a scale hyperparameter to be set up in the section of experiments. The architecture parameters $\textbf{\emph{A}}$ are updated via gradient descent or architecture search algorithms \cite{liu2019darts}, together with the elements of the embedding matrices $\textbf{\emph{W}}$. After the completion of searching, we rank the diagonal elements of the scoring matrix $\textbf{\emph{A}}$ according to their absolute values. The dimensions of the embedding matrices are pruned if their corresponding scores are ranked low. Suppose the remaining ratio of the dimensions after pruning is $\rho$. Then only $\rho d$ dimensions with higher scores are left in the pruned matrices. 

For MSA, we have $\textbf{\emph{W}}_{Q,j}$, $\textbf{\emph{W}}_{K,j}$ and $\textbf{\emph{W}}_{V,j}\in \mathbb{R}^{d\times \rho d/h}$ after pruning, and hence $\textbf{\emph{Q}}_j$,  $\textbf{\emph{K}}_j$ and $\textbf{\emph{V}}_j\in \mathbb{R}^{n\times \rho d/h}$. Since we have not pruned the query or key number $n$, the attention map still belongs to $\mathbb{R}^{n\times n}$, and the head $\textbf{\emph{H}}_j \in \mathbb{R}^{n\times \rho d/h}$. This leads to the projection matrix $\textbf{\emph{W}}_O\in \mathbb{R}^{\rho d\times d}$, and the output of the pruned MSA in $\mathbb{R}^{n\times d}$, with the same shape as the unpruned model. One can easily see that the original unpruned MSA module has $O(4d^2)$ parameters and a computational complexity of $O(4nd^2+2n^2d)$. For the pruned MSA, the number of parameters is reduced to $O(4\rho d^2)$, and the computational complexity is reduced to $O(4\rho nd^2+2\rho n^2d)$. Similarly, the unpruned W-MSA module has $O(4d^2)$ parameters and a computational complexity of $O(4nd^2+2nM^2d)$. For the pruned W-MSA, the number of parameters is reduced to $O(4\rho d^2)$, and the computational complexity is reduced to $O(4\rho nd^2+2\rho nM^2d)$. Finally, the unpruned MLP has $O(2dd_m)$ parameters and a computational complexity of $O(2ndd_m)$. For the pruned MLP, the number of parameters is reduced to $O(2\rho dd_m)$, and the computational complexity is reduced to $O(2\rho ndd_m)$. This is because $\textbf{\emph{W}}_1\in\mathbb{R}^{d\times \rho d_m}$ and $\textbf{\emph{W}}_2\in\mathbb{R}^{\rho d_m\times d}$ after pruning.

One shall note that our settings of architecture parameters are different from those of VTP \cite{zhu2021vision}. VTP's scoring matrix $\textbf{\emph{A}}$ is applied directly to the input feature $\textbf{\emph{X}}$, whereas ours is applied to $\textbf{\emph{Q}}$, $\textbf{\emph{K}}$ and $\textbf{\emph{V}}$. In other words, VTP prunes the features but we prune the linear embeddings. As we apply the same matrix $\textbf{\emph{A}}$ to the embedding dimensions of multiple heads, we have only $d/h$ such architecture parameters, making the model easier to train.  Moreover, VTP is applied to DeiT on the classification task only, whereas our method prunes Swin Transformer, which serves as a backbone for multiple vision tasks. Finally, we have also provided the complexity analysis of the unpruned and pruned operations, which is missing in previous works.

\section{Experiments}

We conduct SiDT for Swin Transformer on CIFAR-100 image classification \cite{krizhevsky2009learning}. We prune its tiny version (Swin-T), which has 27.53M parameters and a complexity of 4.49G FLOPS. The settings of the search stage are similar to those for training the unpruned baseline \footnote{When setting up the architecture parameters, we refer to the code at https://github.com/Cydia2018/ViT-cifar10-pruning}, with batch size = 256, patch size = 4, window size = 7, embedding dimension = 96, initial learning rate = 0.00025, momentum =0.9, weight decay = 0.05, epochs = 160, and the sparsity scale $\gamma = 0.0001$ for $\ell_1$ regularization. After searching, we obtain the scores of all the dimensions and rank them according to their absolute values. Next, the dimensions with lower scores are pruned, based on predefined pruning ratios of 20\%, 40\%, 60\% and 80\%. Finally, the pruned model is trained again with a warm start, using the same settings as the search stage. Table \ref{swin_class} shows that the number of parameters and computational costs can be greatly reduced after pruning, while preserving the accuracy at the same time, compared to the baseline \cite{nested2021}. After pruning 80\% of the dimensions, the accuracy is only around 2\% lower than the recovered baseline. The model with 20\% or 40\% dimensions pruned has an accuracy which is even higher than the baseline model. This can be explained by the relatively larger size of Swin-T on easier datasets like CIFAR, as over-parameterized models can cause overfitting.

Additionally, we have also pruned the Swin-T backbone for the COCO object detection task \cite{lin2014microsoft}, following the settings in the Swin paper \cite{liu2021swin}. That is, batch size = 16, initial learning rate = 0.0001, weight decay = 0.05, epochs = 36, and all the other settings of the backbone are the same as the Swin-T for CIFAR classification discussed above. We use Cascade Mask R-CNN \cite{cai2018cascade} as the detection head, in accordance with that of the Swin-T baseline. Again we follow the steps in Fig. \ref{diag}, and prune the model with pruning ratios of 20\% and 40\%. During the search stage, we also start training with a pretrained Swin-T object detection model. Table \ref{swin_det} indicates that the model with 20\% dimensions of the backbone pruned has a similar performance of box mAP and mask mAP as the unpruned model. Here mAP means the mean average precision over all categories. The box or mask indicates that mAP is computed over bounding boxes or masks. Even if 40\% dimensions of the backbone are pruned, the loss in AP is still less than 1.5\%. This is a fair result since the detection task is more complicated than the classification task, and pruning a detection model can lead to a slightly larger accuracy decline. 

\begin{table}
% \begin{wraptable}{r}{8cm}
    \begin{center}
            \caption{Prune Swin-T via SiDT on CIFAR-100 classification task. PR = Pruning Ratio. Acc = accuracy. Para. = number of parameters. $\diamond$ This baseline is recovered on our device of one RTX 3090 GPU.
    % Non-pretrained models 
    % improve accuracy while gaining a range of sparsities (\%)
    % after channel (ch) pruning. 
    }       \label{swin_class}
    \begin{tabular}{l c c c}
    \toprule
    \specialrule{0em}{1pt}{1pt}
    PR & Acc (\%) & Para. (M) & FLOPS (G)
    % & Prune Rate (\%)
    \\
    \specialrule{0em}{1pt}{1pt}
    \midrule
    \specialrule{0em}{1pt}{1pt}
    0\% (Baseline \cite{nested2021}) &  78.07 & - & -\\
    0\% (Baseline $\diamond$) &  81.78 & 27.60 & 4.49\\
      \specialrule{0em}{1pt}{1pt}
      \midrule
      \specialrule{0em}{1pt}{1pt}
    20\% SiDT & \textbf{82.75} & 23.28 & 3.53 \\
    40\% SiDT & 82.11 & 17.89 & 2.60 \\
    60\% SiDT & 80.81 & 11.92 & 1.73 \\
    80\% SiDT & 79.35 & \textbf{7.17} & \textbf{0.92} \\
    \specialrule{0em}{1pt}{1pt}
    \bottomrule
    \end{tabular}
    \end{center}
\end{table}

\begin{table}
% \begin{wraptable}{r}{8cm}
    \begin{center}
            \caption{Prune Swin-T backbone via SiDT on COCO object detection task. PR = Pruning Ratio.
    % Non-pretrained models 
    % improve accuracy while gaining a range of sparsities (\%)
    % after channel (ch) pruning. 
    }       \label{swin_det}
    \begin{tabular}{l c c   c c}
    \toprule
    \specialrule{0em}{1pt}{1pt}
    \multirow{2}{*}{\makecell{PR}} & 
    \multicolumn{2}{c}{mAP}
     & \multicolumn{2}{c}{Para. (M)}\\
    % & Prune Rate (\%)
    & Box & Mask & Total & Backbone \\
    \specialrule{0em}{1pt}{1pt}
    \midrule
    \specialrule{0em}{1pt}{1pt}
    0\% (Baseline \cite{liu2021swin}) & \textbf{50.5} & \textbf{43.7} & 86 & 28\\
      \specialrule{0em}{1pt}{1pt}
      \midrule
      \specialrule{0em}{1pt}{1pt}
    20\% SiDT & 50.4 & \textbf{43.7} & 80 & 22\\
    40\% SiDT & 49.2 & 42.9 & \textbf{74} & \textbf{16}\\
    \specialrule{0em}{1pt}{1pt}
    \bottomrule
    \end{tabular}
    \end{center}
\end{table}

\section{Conclusion}
We have developed SiDT, a method for searching for the intrinsic dimensions of transformers, and provided its complexity analysis. Experiments on multiple vision tasks have shown that SiDT can promote the efficiency of vision transformers with little accuracy loss. This method will be applied to more computer vision tasks in future work.

\section{Acknowledgements}
The work was partially supported by NSF grants DMS-1854434, DMS-1952644, and a Qualcomm Faculty Award. The authors would like to thank Dr. Shuai Zhang and Dr. Jiancheng Lyu for helpful discussions.

\bibliographystyle{abbrv}
\bibliography{main}
\end{document}